\documentclass{bmvc2k}

\usepackage{booktabs}
\usepackage{tabularx}
\usepackage{arydshln}
\newcommand{\ra}[1]{\renewcommand{\arraystretch}{#1}}
\newcolumntype{Y}{>{\centering\arraybackslash}X}

\title{Automated Search for Resource-Efficient Branched Multi-Task Networks}

\addauthor{David Bruggemann}{brdavid@vision.ee.ethz.ch}{1}
\addauthor{Menelaos Kanakis}{kanakism@vision.ee.ethz.ch}{1}
\addauthor{Stamatios Georgoulis}{georgous@vision.ee.ethz.ch}{1}
\addauthor{Luc Van Gool}{vangool@vision.ee.ethz.ch}{1}

\addinstitution{
 Computer Vision Lab\\
 ETH Zurich\\
 Switzerland
}

\runninghead{Bruggemann et al.}{Branched Multi-Task Networks}

\def\eg{e.g\bmvaOneDot}

\newcommand{\beginappendixa}{%
        \setcounter{table}{0}
        \renewcommand{\thetable}{A-\arabic{table}}%
        \setcounter{figure}{0}
        \renewcommand{\thefigure}{A-\arabic{figure}}%
     }
     
\newcommand{\beginappendixb}{%
        \setcounter{table}{0}
        \renewcommand{\thetable}{B-\arabic{table}}%
        \setcounter{figure}{0}
        \renewcommand{\thefigure}{B-\arabic{figure}}%
     }
\newcommand{\beginappendixc}{%
        \setcounter{table}{0}
        \renewcommand{\thetable}{C-\arabic{table}}%
        \setcounter{figure}{0}
        \renewcommand{\thefigure}{C-\arabic{figure}}%
     }
\newcommand{\beginappendixd}{%
        \setcounter{table}{0}
        \renewcommand{\thetable}{D-\arabic{table}}%
        \setcounter{figure}{0}
        \renewcommand{\thefigure}{D-\arabic{figure}}%
     }
     
\usepackage[page]{appendix}

\begin{document}

\maketitle

\begin{abstract}
\looseness=-1
The multi-modal nature of many vision problems calls for neural network architectures that can perform multiple tasks concurrently. Typically, such architectures have been handcrafted in the literature. However, given the size and complexity of the problem, this manual architecture exploration likely exceeds human design abilities. In this paper, we propose a principled approach, rooted in differentiable neural architecture search, to automatically define branching (tree-like) structures in the encoding stage of a multi-task neural network. To allow flexibility within resource-constrained environments, we introduce a proxyless, resource-aware loss that dynamically controls the model size. Evaluations across a variety of dense prediction tasks show that our approach consistently finds high-performing branching structures within limited resource budgets.
\end{abstract}
\section{Introduction}

Over the last decade neural networks have shown impressive results for a multitude of tasks. This is typically achieved by designing network architectures that satisfy the specific needs of the task at hand, and training them in a fully supervised fashion~\cite{girshick2014rich,long2015fully,xie2015holistically}. This well-established strategy assumes that each task is tackled in isolation, and consequently, a dedicated network should be learned for every task. However, real-world problems are inherently multi-modal (\eg, an autonomous car should be able to detect road lanes, recognize pedestrians in its vicinity, semantically understand its surroundings, estimate its distance from other objects, etc.), which calls for architectures that can perform multiple tasks simultaneously. 

Motivated by these observations researchers started designing multi-task architectures that, given an input image, can produce predictions for all desired tasks~\cite{ruder2017overview}. Arguably the most characteristic example is branched multi-task networks~\cite{kokkinos2017ubernet,neven2017fast,long2017learning,kendall2018multi,chen2017gradnorm}, where a shared encoding stage branches out to a set of task-specific heads that decode the shared features to yield task predictions. Follow-up works~\cite{misra2016cross,xu2018pad,liu2019end,gao2019nddr,maninis2019attentive,vandenhende2020mti} proposed more advanced mechanisms for multi-task learning. However, despite their advantages, they similarly assume that the multi-task architecture can be handcrafted prior to training. In a different vein, some works~\cite{lu2017fully,vandenhende2019branched} opted for a semi-automated architecture design where the branching occurs at finer locations in the encoding stage. Task groupings at each branching location are determined based on a measure of `task relatedness' in pre-trained networks. However, defining branching points based on such offline criteria disregards potential optimization benefits resulting from jointly learning particular groups of tasks~\cite{bingel2017identifying}.

Generally, finding an architecture suitable for multi-task learning poses great challenges, arising from conflicting objectives. On the one hand, a multi-task network should perform comparably to its single-task counterparts. This is not trivial as task interference\footnote{Task interference is a well-documented problem in multi-task networks~\cite{kokkinos2017ubernet,kendall2018multi,sener2018multi,maninis2019attentive} where important information for one task might be a nuisance for another, leading to conflicts in the optimization of shared parameters.} can significantly affect individual task performance. On the other hand, a multi-task network should remain within a low computational budget during inference with respect to the single-task case. The relative importance of task performance vs.\ computational efficiency depends on the application, highlighting the necessity of being able to adapt the architecture design flexibly. This calls for automatic architecture search techniques to mitigate the effort accompanying architecture handcrafting. 

In this paper, we address the aforementioned needs and propose Branched Multi-Task Architecture Search (BMTAS\footnote{Reference code at \url{https://github.com/brdav/bmtas}}), a principled approach to automatically determine encoder branching in multi-task architectures. To avoid a brute-force or greedy search as in~\cite{vandenhende2019branched} and~\cite{lu2017fully} respectively, we build a differentiable neural architecture search algorithm with a search space directly encompassing all possible branching structures. Our approach is end-to-end trainable, and allows flexibility in the model construction through the introduction of a proxyless, resource-aware loss. Experimental analysis on multiple dense prediction tasks shows that models generated by our method effectively balance the trade-off between overall performance and computational cost.
\section{Related Work}

\textbf{Multi-Task Learning} (MTL) in deep neural networks addresses the problem of training a single model that can perform multiple tasks concurrently. To this end, a first group of works~\cite{misra2016cross,ruder2019latent,liu2019end,gao2019nddr} incorporated feature sharing mechanisms on top of a set of task-specific networks. Typically, these \textit{soft parameter sharing} approaches scale poorly to an increasing number of tasks, as the parameter count usually exceeds the single-task case. A second group of works~\cite{kokkinos2017ubernet,neven2017fast,long2017learning,kendall2018multi,chen2017gradnorm} shared the majority of network operations, before branching out to a set of task-specific heads that produce the task predictions. In these \textit{hard parameter sharing} approaches the branching point is manually determined prior to training, which can lead to task interference if `unrelated' (groups of) tasks are selected.

Follow-up works employed different techniques to address the MTL problem. PAD-Net~\cite{xu2018pad} and MTI-Net~\cite{vandenhende2020mti} proposed to refine the initial task predictions by distilling cross-task information. Dynamic task prioritization~\cite{guo2018dynamic} constructed a hierarchical network that dynamically prioritizes the learning of `easy' tasks at earlier layers. ASTMT~\cite{maninis2019attentive} and RCM~\cite{kanakis2020reparameterizing} adopted a task-conditional approach, where only one task is forward-propagated at a time, to sequentially generate all task predictions. All the above mentioned approaches assume that the multi-task architecture can be handcrafted. However, given the size and complexity of the problem, this manual exploration likely exceeds human design abilities. In contrast, stochastic filter groups~\cite{bragman2019stochastic} re-purposed the convolution kernels in each layer to support shared or task-specific behavior, but their method only operates at the channel level of the network.  

Our work is closer to~\cite{lu2017fully,vandenhende2019branched} where branching structures are generated in the encoding stage of the network. However, instead of determining the task groupings at each branching location in pre-trained networks via a brute-force or greedy algorithm, we propose a principled solution rooted in differentiable neural architecture search that is trainable end-to-end.

\textbf{Neural Architecture Search} (NAS) aims to automate the procedure of designing neural network architectures in contrast to the established protocol of manually drafting them based on prior knowledge. To achieve this, a first group of works~\cite{zoph2016neural,zoph2018learning,pham2018efficient} used a recurrent neural network to sample architectures from a pre-defined search space and trained it with reinforcement learning to maximize the expected accuracy on a validation set. A second group of works~\cite{real2017large,real2019regularized} employed evolutionary algorithms to gradually evolve a population of models through mutations. Despite their great success, these works are computationally demanding during the search phase, which motivated researchers to explore differentiable NAS~\cite{liu2018darts} through continuous relaxation of the architecture representation.
SNAS~\cite{xie2018snas} built upon Gumbel-Softmax~\cite{jang2016categorical,maddison2016concrete} to propose an effective search gradient.
Follow-up works~\cite{cai2018proxylessnas,wu2019fbnet} introduced resource constraints into the optimization pipeline to control the model size. 

In general, NAS works have mainly focused on the image classification task, with a few exceptions that addressed other tasks too (\eg, semantic segmentation~\cite{liu2019auto}). Moreover, each task is tackled in isolation, i.e., a dedicated architecture is generated for each task. When it comes to MTL however, there is a need to design architectures that perform well across a variety of tasks. Routing networks~\cite{rosenbaum2017routing} made a first attempt in this direction by determining the connectivity of a network's function blocks through routing, but they focus on classification tasks, as opposed to the more challenging dense prediction considered in this paper. MTL-NAS~\cite{gao2020mtl} extended NDDR-CNN~\cite{gao2019nddr} to automatically find the feature fusion mechanisms on top of the task-specific networks, but their search is solely limited to the feature sharing mechanisms. In contrast, we automatically generate branching (or tree-like) structures in the encoding stage of the network, which allows more explicit control over the size of the final model. 

\section{Method}

Given a set of dense prediction tasks and an arbitrary neural architecture, our goal is to find resource-efficient branching structures in the encoder, that promote the sharing of general-purpose features between tasks and the decoupling of task-specific features across tasks. In this section, we elaborate on three key components of the proposed BMTAS, which builds upon differentiable NAS to tackle the aforementioned objective: the structure of the search space (Sec.~\ref{subsec:branch}), the algorithm to traverse that search space (Sec.~\ref{subsec:algo}), and a novel objective function to enforce resource efficiency (Sec.~\ref{subsec:res_loss}).

\subsection{Search Space}
\label{subsec:branch}

\begin{figure}[t]
\includegraphics[width=\textwidth]{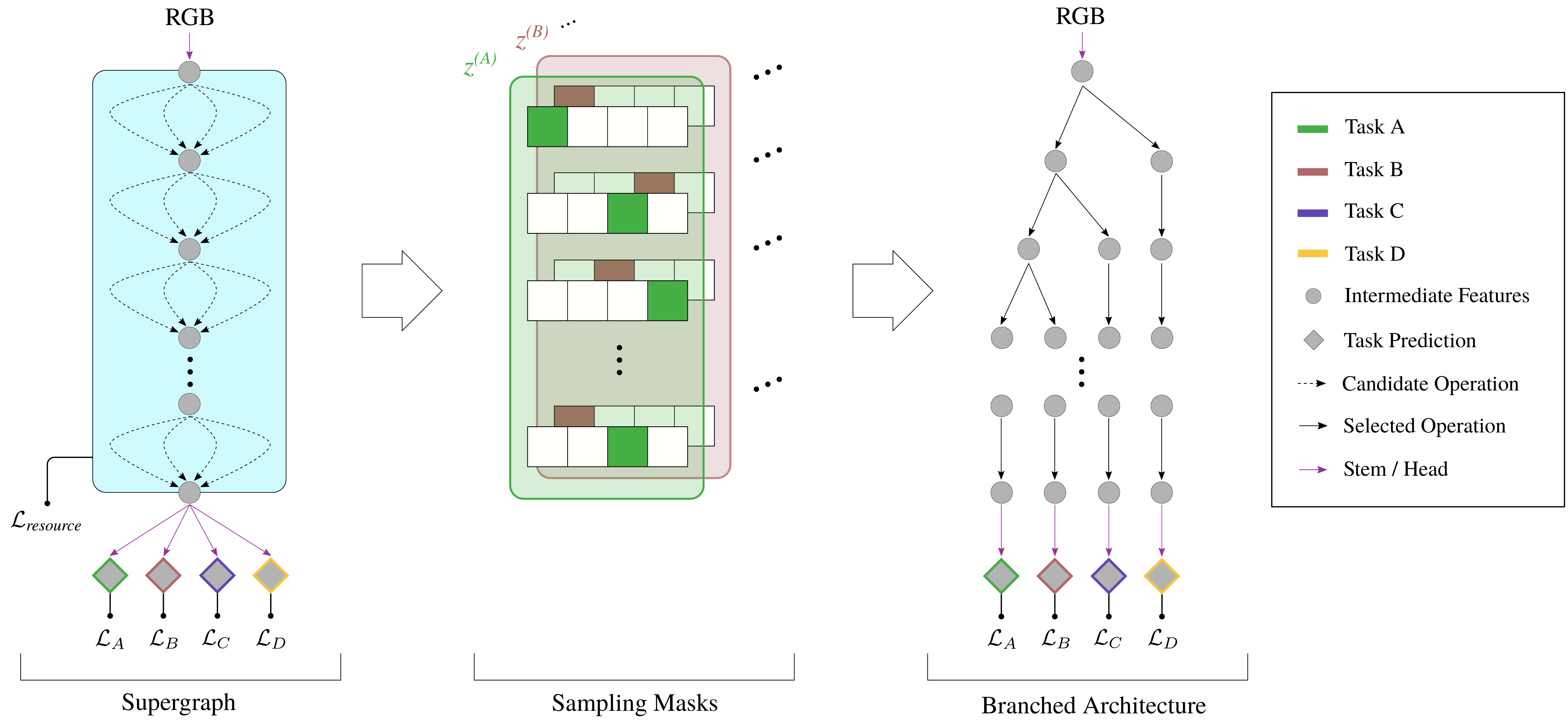}
\caption{Schematic showing the construction of a branched architecture (right) from a supergraph (left), in a setting with four tasks. For each of the tasks A to D, a subgraph is sampled from the supergraph, using the learnable masks $z^{(\cdot)}$. To form a branching structure, the subgraphs are combined according to the sampling consensus, yielding task groupings at each layer. During the architecture search, the masks $z^{(\cdot)}$ are learned by minimizing a resource loss $\mathcal{L}_{resource}$ (computed using a look-up table) and task performance losses $\mathcal{L}_A$ to $\mathcal{L}_D$ simultaneously.}
\label{fig:method}
\end{figure}

In contrast to most established NAS works~\cite{zoph2016neural,zoph2018learning,pham2018efficient,liu2018darts} which only search for subcomponents (\eg, a cell) of the target architecture, our search space directly encompasses all possible branching structures for a given number of $T$ tasks in an encoder network. As shown in Fig.~\ref{fig:method}, we can describe our search space with a directed acyclic graph (cyan box), where the vertices represent intermediate feature tensors and the edges operations (\eg, bottleneck blocks for a ResNet-50 backbone). For an encoder with $L$ layers the graph has length $L$ in total, and width $T$ between consecutive vertices. Parallel edges denote candidate operations, which are (non-parameter-sharing) duplicates of the original operation in the respective layer. The operation parameters of each duplicate block are `warmed up' for a few iterations on the corresponding task before the architecture search. Through this procedure, each block is softly assigned to a task. For any one task $t$, a specific routing (subgraph) through the supergraph can be obtained by sampling operations with a mask $z^{(t)} \in \{0, 1\}^{L \times T}$ with one-hot rows (Fig.~\ref{fig:method} center). Any possible branching structure for a set of tasks can then be produced by combining the task-specific routings. 
Importantly, computation sharing in layer $l$ for any two tasks only occurs if their sampled edges coincide in all layers 1 to $l$.
We present a loss function which incentivizes such computation sharing in Sec.~\ref{subsec:res_loss}.

\subsection{Search Algorithm}
\label{subsec:algo}

The goal of BMTAS is to find a set of masks $Z$ = $\{z^{(t)} | t \in \{1, ..., T\}\}$ to sample $T$ subgraphs from the supergraph described in Sec.~\ref{subsec:branch}. We model this set with a parameterized distribution $p_\alpha(Z)$, where $\alpha^{(t)} \in \mathbb{R}^{L\times T}$ represents the unnormalized log probabilities for sampling individual operations for task $t$. Therefore, the overall optimization problem is:
\begin{equation}
    \min_{\alpha, \theta} \mathbb{E}_{Z\sim p_\alpha(Z)} \left[\mathcal{L}\left(Z, \theta\right)\right]
    \label{eq:perf}
\end{equation}
$\theta$ denotes the neural operation parameters of the supergraph and $\mathcal{L}$ an appropriate loss function (see Sec.~\ref{subsec:res_loss}). By solving this optimization problem, we are maximizing the expected performance (in accordance with the chosen $\mathcal{L}$) of branching structures sampled from $p_\alpha(Z)$. 

Following~\cite{xie2018snas}, we can relax the discrete architecture distribution $p_\alpha(Z)$ to be continuous and differentiable using the gradient estimator proposed in~\cite{jang2016categorical,maddison2016concrete}: 
\begin{equation}
    z_{l,j}^{(t)} = \frac{\exp{\left(\left(\alpha_{l,j}^{(t)} + g_{l,j}^{(t)}\right)/\tau\right)}}{\sum_{i=1}^T \exp{\left(\left(\alpha_{l,i}^{(t)} + g_{l,i}^{(t)}\right)/\tau\right)}}, \quad j=1, ..., T
\end{equation}
$g_{l,j}^{(t)} \sim \text{{\tt Gumbel}}(0, 1)$ is random noise and $\tau>0$ a temperature parameter. If $\tau$ is very large, the sampling distribution is nearly uniform, regardless of the $\alpha^{(t)}$ values. This has the advantage that the resulting gradients in the backpropagation are smooth. For $\tau$ close to 0, we approach sampling from the categorical distribution $p_\alpha(Z)$. In practice, we gradually anneal $\tau \rightarrow 0$ during training. Samples from the supergraph are obtained by multiplying the edges in layer $l$ with the softened one-hot vector $z_l^{(t)}$ and summing the output.

The above function, known as Gumbel-Softmax, enables us to directly learn the unnormalized log probability masks $\alpha^{(t)}$ through gradient descent for each task $t$. During training, we alternate between updating the architecture parameters $\alpha$ and the operation parameters $\theta$, as this empirically leads to more stable convergence in our case. Afterwards, we discretize $\alpha^{(t)}$ using {\tt argmax} to obtain one-hot masks that determine the final routing. As is common practice~\cite{liu2018darts, xie2018snas}, we retrain the searched architectures from scratch.

\subsection{Resource-Aware Objective Function}
\label{subsec:res_loss}

In the simplest case, the loss function in the overall optimization problem (Eq.~\ref{eq:perf}) of BMTAS consists of a weighted sum of the task-specific losses:
\begin{equation}
    \mathcal{L}_{tasks}(Z, \theta) = \sum_{t=1}^T \omega_t \mathcal{L}_t(Z, \theta)
\label{eq:weighted_sum}
\end{equation}
Simply searching encoder structures via this objective function yields performance-oriented branching structures, irrespective of the efficiency of the resulting model. We show in Sec.~\ref{subsec:ablations} that the outcome resembles separate single-task models, i.e., the tasks stop sharing computations and branch out in early layers, increasing the capacity of the resulting network. To obtain more compact encoder structures and to actively navigate the efficiency vs.\ performance trade-off, we introduce a resource-aware term $\mathcal{L}_{resource}$ in the objective function:
\begin{equation}
    \mathcal{L}_{search} = \mathcal{L}_{tasks} + \lambda\mathcal{L}_{resource}
\label{eq:tot_loss}
\end{equation}
The emphasis shifts to resource efficiency in the architecture search as $\lambda$ is increased. We follow related work~\cite{gordon2018morphnet,molchanov2016pruning,veniat2018learning} in choosing the number of multiply-add operations (MAdds) during inference as a surrogate for resource efficiency. The MAdds $C(Z)$ of a sampled branched architecture $Z$ depends on the present \textit{task groupings} in each network layer. We define a task grouping as a \textit{partition} of the set $\{1, ..., T\}$, where the parts indicate computation sharing. For an encoder with $L$ layers and $K$ possible groupings per layer, the resource objective can be formulated as:
\begin{equation}
    \mathbb{E}_{Z\sim p_\alpha(Z)}\left[C\left(Z\right)\right] = \sum_{l=1}^{L}\sum_{k=1}^{K} p_\alpha(\kappa_l = k) c(k, l)
\label{eq:res_decomp}
\end{equation}
where $\kappa_l$ is the task grouping at layer $l$, and $c(k, l)$ the MAdds of grouping $k$ at layer $l$. $c(k, l)$ can be simply determined by a look-up table, as the computational cost only depends on the layer index and the number of required operations given the grouping. 

In contrast to differentiable NAS works~\cite{xie2018snas,wu2019fbnet,cai2018proxylessnas}, we cannot calculate \mbox{$p_\alpha(\kappa_l = k)$} for each layer independently, since task groupings depend on previous layers. As mentioned in Sec.~\ref{subsec:branch}, two tasks only share computation in layer $l$ provided that they also do so in all preceding layers.
Thus, for each task grouping $k$, we need to map out the set of valid ancestor groupings $\mathcal{A}_k$. In mathematical terms, $\mathcal{A}_k$ contains every partition (i.e., task grouping) of which $k$ is a \textit{refinement}. Computation sharing with grouping $k$ in layer $l$ only occurs if the groupings in every layer 1 to $(l-1)$ are in $\mathcal{A}_k$. Considering this dependency structure, we can decompose \mbox{$p_\alpha(\kappa_l = k)$} as a recursive formulation of conditional probabilities:
\begin{equation}
    p_\alpha(\kappa_l=k)=p_\alpha(\kappa_l=k | \kappa_{l-1}, ..., \kappa_1 \in\mathcal{A}_k)\sum_{m\in\mathcal{A}_k} p_\alpha(\kappa_{l-1}=m), \quad l=2, ..., L
\label{eq:res_rec}
\end{equation}
The conditional probabilities $p_\alpha(\kappa_l=k | \kappa_{l-1}, ..., \kappa_1 \in\mathcal{A}_k)$ are independent for each layer and can be easily constructed from the unnormalized log sampling probabilities $\alpha$.

The derivations presented in this section yield a \textit{proxyless} resource loss function, i.e., it encourages solutions which directly minimize the expected resource cost of the final model. By design, this resource objective function allows us to find tree-like structures without resorting to curriculum learning techniques during the architecture search, which would be infeasible using simpler, indirect constraints (\eg, $\mathcal{L}_2$ regularization).

\section{Experiments}
\label{sec:experiments}

In this section, we first describe the experimental setup (Sec.~\ref{subsec:expsetup}), and consequently evaluate the proposed method on several datasets using different backbones (Sec.~\ref{subsec:results}). We also present ablation studies to validate our search algorithm (Sec.~\ref{subsec:ablations}), and analyze the resulting task groupings in a case study (Sec.~\ref{subsec:taskpair}).

\subsection{Experimental Setup}
\label{subsec:expsetup}

\hspace{\parindent}\textbf{Datasets.} We carry out experiments on PASCAL~\cite{everingham2010pascal} and NYUD-v2~\cite{silberman2012indoor}, two popular datasets for dense prediction MTL. 
For PASCAL we use the PASCAL-Context~\cite{chen2014detect} data split, which includes 4998 training and 5105 testing images, densely labeled for semantic segmentation (SemSeg), human parts segmentation (PartSeg), saliency estimation (Sal), surface normal estimation (Norm), and edge detection (Edge). We adopt the distilled saliency and surface normal labels from~\cite{maninis2019attentive}.
The NYUD-v2 dataset comprises 795 training and 654 testing images of indoor scenes, fully labeled for semantic segmentation (SemSeg), depth estimation (Depth), surface normal estimation (Norm), and edge detection (Edge).
All training and evaluation of the baselines and final branched models was conducted on the full-resolution images. Yet, to accelerate the architecture search, we optimize BMTAS using resized input images (1/2 and 2/3 resolution for PASCAL-Context and NYUD-v2 respectively).

\textbf{Architectures.} For all experiments we use a DeepLabv3+ base architecture~\cite{chen2018encoder}, which was designed for semantic segmentation, but was shown to perform well for various dense prediction tasks~\cite{maninis2019attentive,kanakis2020reparameterizing}. To demonstrate generalization of our method across encoder types, we report results for both MobileNetV2~\cite{sandler2018mobilenetv2} and ResNet-50~\cite{he2016deep} backbones. For MobileNetV2 we employ a reduced design of the ASPP module (R-ASPP), proposed in~\cite{sandler2018mobilenetv2}.

\textbf{Metrics.} We evaluate semantic segmentation, human parts segmentation and saliency estimation using mean intersection over union. For surface normal estimation we use mean angular error, for edge detection optimal dataset F-measure~\cite{martin2004learning}, and for depth estimation root mean square error. To obtain a single-number performance metric for a multi-task model, we additionally report the average per-task performance drop ($\Delta_m$) with respect to single-task baselines~\textit{b} for model~\textit{m}, as proposed in~\cite{maninis2019attentive}. It is defined as 
\begin{equation}
\Delta_m = \frac{1}{T} \sum_{i=1}^{T} (-1)^{l_i} \left(M_{m,i} - M_{b,i}\right) / M_{b,i}
\end{equation}
where $l_i = 1$ if a lower value for metric $M_i$ indicates better performance, and $l_i = 0$ otherwise.

\textbf{Baselines.} To eliminate the influence of differences in training setup (data augmentation, hyperparameters, etc.), we compare BMTAS with our own implementations of the baselines. As a reference, we first report the performance of standard single-task models (`Single') and a multi-task model consisting of a fully shared encoder and task-specific heads (`Shared'). We consider the ASPP (resp.\ R-ASPP) module part of the encoder, and as such, share it between all tasks for this model. 
Furthermore, we create branched encoder structures by two competing methods: \textit{i)} In FAFS~\cite{lu2017fully}, a fully shared encoder is first pre-trained and then greedily split layer by layer, starting from the final layer. Branches are separated based on task similarity in sample space, taking into account the complexity of the resulting model. \textit{ii)} In BMN~\cite{vandenhende2019branched}, branching structures are found by measuring task affinity with feature map correlations between pre-trained single-task networks. The final models are the result of an exhaustive search over all possible branching configurations, capped at maximum resource cost. Finally, we evaluate the performances of three state-of-the-art MTL approaches: Cross-Stitch Networks (C.-S.)~\cite{misra2016cross}, NDDR-CNN~\cite{gao2019nddr} and MTAN~\cite{liu2019end}.

\subsection{Comparison with the State-of-the-Art}
\label{subsec:results}

\begin{figure}
\includegraphics[width=0.95\textwidth]{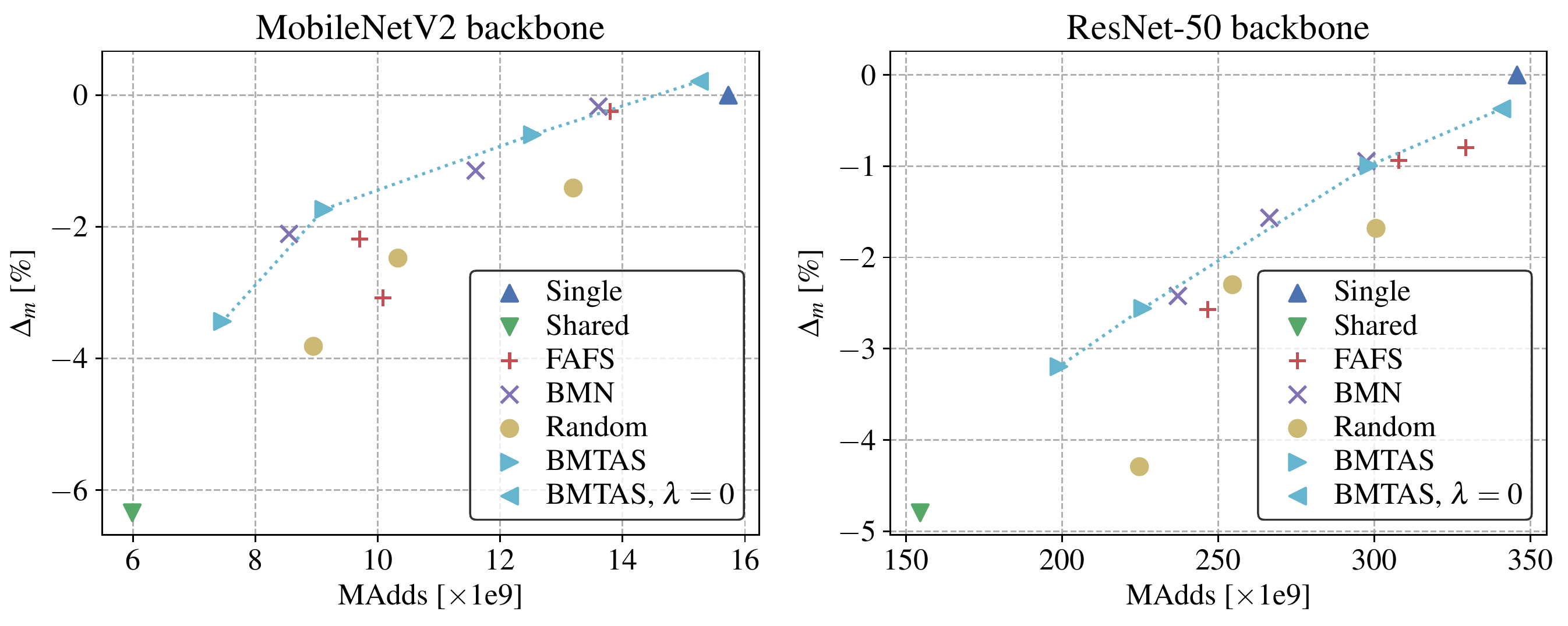} 
\caption{Comparison of multi-task performance $\Delta_m$ as a function of multiply-add operations (MAdds) of encoder branching methods on PASCAL-Context, for a MobileNetV2 backbone (left) and a ResNet-50 backbone (right). For all methods three models with adjustable complexity penalty were evaluated---except ours (BMTAS), for which we additionally report the model obtained by an architecture search without resource loss (no complexity penalty). BMTAS models favorably balance the efficiency vs.\ performance trade-off.}
\label{fig:res}
\end{figure}

\begin{table}
\begin{center}
\resizebox{\linewidth}{!}{%
\ra{1.3}
\footnotesize
\begin{tabularx}{\textwidth}{@{}lYYYYYYr@{}}\toprule
Model & MAdds\,$\downarrow$ & SemSeg\,$\uparrow$ & PartSeg\,$\uparrow$ & Sal\,$\uparrow$ & Norm\,$\downarrow$ & Edge\,$\uparrow$ & $\Delta_m$ [\%]\,$\uparrow$ \\ \midrule
Single & 15.7B & 65.11 & 57.54 & 65.41 & 13.98 & 69.50 & 0.00 \\
Shared & \phantom{0}6.0B & 59.69 & 55.96 & 63.03 & 16.02 & 67.80 & -6.35 \\
\hdashline\noalign{\vskip 0.5ex}
C.-S.~\cite{misra2016cross} & 15.7B & 63.28 & 60.21 & 65.13 & 14.17 & 71.10 & 0.47 \\
NDDR-CNN~\cite{gao2019nddr} & 21.8B & 63.22 & 56.12 & 65.16 & 14.47 & 68.90 & -2.02 \\
MTAN~\cite{liu2019end} & \phantom{0}9.5B & 61.55 & 58.89 & 64.96 & 14.74 & 69.90 & -1.73 \\
\hdashline\noalign{\vskip 0.5ex}
BMTAS-1 & \phantom{0}7.5B & 61.43 & 56.77 & 63.64 & 14.77 & 68.20 & -3.44 \\
BMTAS-2 & \phantom{0}9.1B & 62.80 & 57.72 & 64.92 & 14.48 & 68.70 & -1.74 \\
BMTAS-3 & 12.5B & 64.07 & 58.60 & 64.72 & 14.27 & 69.40 & -0.60 \\
\bottomrule
\end{tabularx}
}
\end{center}
\caption{Comparison of BMTAS with MTL baselines on PASCAL-Context using a MobileNetV2 backbone. The resource loss weights $\lambda$ for BMTAS-\{1, 2, 3\} are chosen on a logarithmic scale: \{0.1, 0.05, 0.02\} respectively.}
\label{tab:mobile_pascal}
\end{table}

Since the proposed method is targeted towards resource-constrained applications, we optimize two objectives concurrently: multi-task performance and resource cost, and we ultimately seek pareto-efficient solutions. Fig.~\ref{fig:res} depicts a comparison of the principal encoder branching methods. For each method, three models are generated by varying the complexity penalties in the algorithms. Our models compare favorably to random structures and FAFS, and perform on par with BMN. Unlike BMN however, our approach is end-to-end trainable, and does not rely on an offline brute-force search over the entire space of configurations.

Table~\ref{tab:mobile_pascal} and Table~\ref{tab:resnet_nyud} show a breakdown of individual task metrics for our method and state-of-the-art MTL approaches on PASCAL-Context and NYUD-v2 using a MobileNetV2 and ResNet-50 backbone, respectively. On PASCAL-Context, the BMTAS multi-task performances lie between the `Shared' and `Single' baselines. Although the C.-S.\ network performs best in this setting, it also requires considerably more MAdds than all the BMTAS models. On the smaller NYUD-v2 datasets, MTL approaches generally perform better compared to the `Single' baseline. Owing to the large number of channels in the feature maps, both NDDR-CNN and MTAN scale inefficiently to a ResNet-50 backbone. The performance boost for these methods therefore comes at the cost of a sizeable increase in resource cost.

\begin{table}
\begin{center}
\resizebox{\linewidth}{!}{%
\ra{1.3}
\footnotesize
\begin{tabularx}{\textwidth}{@{}lYYYYYr@{}}\toprule
Model & MAdds\,$\downarrow$ & SemSeg\,$\uparrow$ & Depth\,$\downarrow$ & Norm\,$\downarrow$ & Edge\,$\uparrow$ & $\Delta_m$ [\%]\,$\uparrow$ \\ \midrule
Single & 254.4B & 40.08 & 0.5479 & 21.67 & 70.10 & 0.00 \\
Shared & 122.2B & 38.37 & 0.5766 & 22.66 & 70.90 & -3.23 \\
\hdashline\noalign{\vskip 0.5ex}
C.-S.~\cite{misra2016cross} & 254.4B & 41.01 & 0.5380 & 22.00 & 70.40 & 0.77 \\
NDDR-CNN~\cite{gao2019nddr} & 366.2B & 40.88 & 0.5358 & 21.86 & 70.30 & 0.91 \\
MTAN~\cite{liu2019end} & 600.1B & 42.03 & 0.5191 & 21.89 & 70.40 & 2.38 \\
\hdashline\noalign{\vskip 0.5ex}
BMTAS-1 & 155.8B & 40.66 & 0.5691 & 21.84 & 70.70 & -0.58 \\
BMTAS-2 & 223.7B & 40.37 & 0.5413 & 21.74 & 69.80 & 0.30 \\
BMTAS-3 & 248.3B & 41.10 & 0.5431 & 21.56 & 70.10 & 0.98 \\
\bottomrule
\end{tabularx}
}
\end{center}
\caption{Comparison of BMTAS with MTL baselines on NYUD-v2 using a ResNet-50 backbone. The resource loss weights $\lambda$ for BMTAS are \{0.005, 0.001, 0.0002\} respectively.}
\label{tab:resnet_nyud}
\end{table}

Overall, BMTAS models gain the advantage in applications where both resource efficiency and task performance are essential. The method is universally applicable to any backbone and thus able to produce compact multi-task models in all tested scenarios. Furthermore, BMTAS is freely adaptable to the application-specific resource budget, a flexibility which is not provided by C.-S., NDDR-CNN or MTAN without changing the backbone. However, the proposed approach also shows some limitations. Notably, the number of possible task groupings grows quickly with increasing tasks\footnote{The total number of task groupings for $T$ tasks is given by the Bell number $B_T$.}, raising the complexity of computing the resource loss. Combined with the enlarged space of possible branching configurations, this can slow down the architecture search considerably for many-task learning. 

For a medium number of tasks however, BMTAS is reasonably efficient. Concretely, the search time was around 1 day on NYUD-v2 (4 tasks, 25000 iterations) and 3.5 days on PASCAL-Context (5 tasks, 50000 iterations) for either backbone using a single V100 GPU.

\subsection{Ablation Studies}
\label{subsec:ablations}

In Table~\ref{tab:ablation} we present two sets of ablation studies on our search algorithm. As a reference model, we use `BMTAS-2' from Table~\ref{tab:mobile_pascal}. 

First, we validate the efficacy of the obtained branching by randomly permuting task-to-branch pairings in the reference structure, before training it (`Permutations'). We repeat this procedure five times and report the mean scores and standard deviations in Table~\ref{tab:ablation}. We also train a simple branching structure consisting of some fully shared layers and simultaneous branching for all tasks (`Vanilla'). The branching point is chosen such that the number of MAdds coincides with our reference model. Although these two baselines outperform our reference model on SemSeg and Edge, the overall multi-task performance is lower in both cases, suggesting that our method learns to disentangle tasks more efficiently.

\begin{table}
\begin{center}
\resizebox{\linewidth}{!}{%
\ra{1.3}
\footnotesize
\begin{tabularx}{\textwidth}{@{}lYYYYYr@{}}\toprule
Model & SemSeg $\uparrow$ & PartSeg $\uparrow$ & Sal $\uparrow$ & Norm $\downarrow$ & Edge $\uparrow$ & $\Delta_m$ [\%] $\uparrow$ \\ \midrule
BMTAS & 62.80 & 57.72 & 64.92 & 14.48 & 68.70 & -1.74 \\
\hdashline\noalign{\vskip 0.5ex}
Permutations & 63.07\tiny{$\pm$1.09} & 57.46\tiny{$\pm$0.39} & 64.25\tiny{$\pm$0.20} & 14.94\tiny{$\pm$0.25} & 69.20\tiny{$\pm$0.73} & -2.47\tiny{$\pm$0.39} \\
Vanilla & 64.29 & 57.15 & 64.43 & 15.06 & 69.10 & -2.34 \\
\hdashline\noalign{\vskip 0.5ex}
w/o res.\ loss & 64.70 & 58.82 & 65.32 & 14.14 & 70.00 & 0.21 \\
w/o warm-up & 62.70 & 57.93 & 63.66 & 14.74 & 68.60 & -2.49 \\
w/o resizing & 63.29 & 57.50 & 64.06 & 14.37 & 68.70 & -1.77 \\
\bottomrule
\end{tabularx}
}
\end{center}
\caption{Ablation studies on PASCAL-Context using a MobileNetV2 backbone. The numbers for `Permutations' reflect the mean and standard deviation of five independent runs.}
\label{tab:ablation}
\end{table}

Second, we ablate three components of our search algorithm and report the corresponding results in Table~\ref{tab:ablation}: \textit{i)}~We discard the resource term in the loss (`w/o res.\ loss'), i.e., we set $\lambda=0$ in the total search loss of Eq.~\ref{eq:tot_loss}. Both the performance and resource cost (see also Fig.~\ref{fig:res} left) of the found model are close to the single-task case, as suggested in Sec.~\ref{subsec:res_loss}. \textit{ii)}~We disable the task-specific `warm-up' of the parallel candidate operations (`w/o warm-up', see Sec.~\ref{subsec:branch}) and directly search architectures with ImageNet initialization for $\theta$. The resulting model underperforms our reference model, indicating the importance of softly assigning each candidate operation to a particular task before starting the search: This helps counteract the bias of the algorithm toward full operation sharing early in the training, as well as breaks parameter initialization symmetry. Only the neural operation parameters $\theta$ are `warmed up' before the search, while the architecture distribution parameters $\alpha$ are simply initialized with zeros. \textit{iii)}~We conduct the architecture search on the full resolution input images (`w/o resizing'), instead of resizing them as mentioned in Sec.~\ref{subsec:expsetup}. The performance remains approximately equal to the reference model, justifying the use of resized images during architecture search.

\subsection{Task Pairing Case Study}
\label{subsec:taskpair}

\begin{figure}
\includegraphics[width=\textwidth]{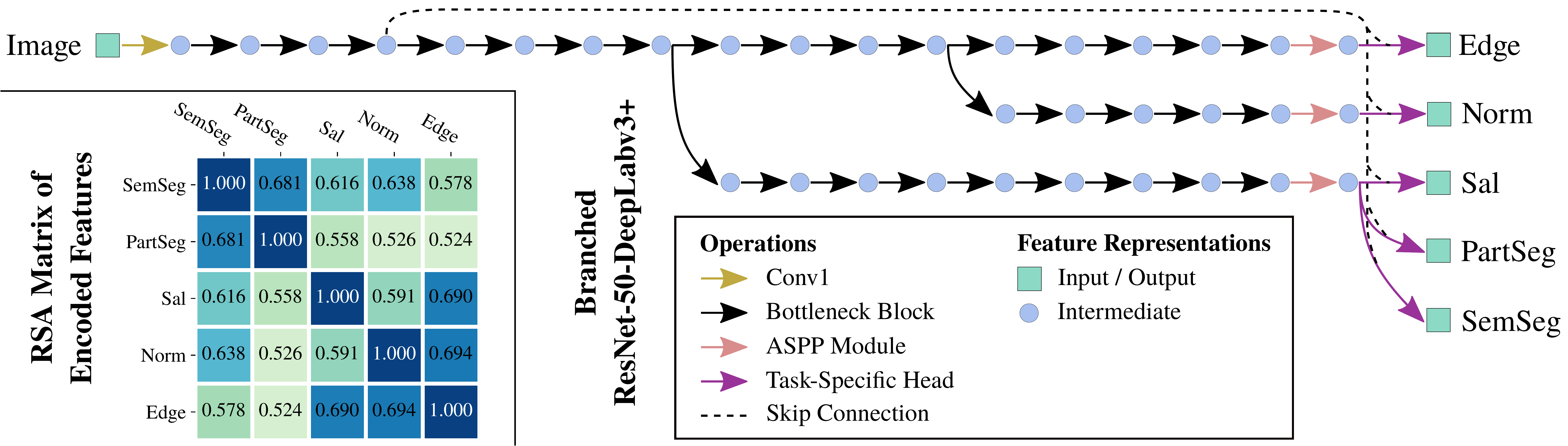}
\caption{Graph of a ResNet-50-DeepLabv3+ branching structure obtained with our method on the PASCAL-Context dataset (`BMTAS-2' in Table~\ref{tab:resnet_pascal}). Edges of the graph indicate operations and vertices indicate feature tensors. On the left, a Representational Similarity Analysis (RSA) matrix determined from correlating encoder output feature maps of the single-task networks is shown. Grouped tasks in the branched architecture exhibit high RSA correlations, validating our searched configuration.}
\label{fig:pair}
\end{figure}

Fig.~\ref{fig:pair} shows a sample ResNet-50 encoder branching structure determined with our method. We compare the resulting task groupings with the Representational Similarity Analysis (RSA) matrix obtained from correlating encoded feature maps of the individual single-task networks, as proposed in~\cite{vandenhende2019branched}. The RSA suggests task groupings similar to ours, with the pairs SemSeg-PartSeg and Norm-Edge having high task affinities in both cases.

\section{Conclusion}

We presented BMTAS, an end-to-end approach for automatically finding efficient encoder branching structures for MTL. Individual task routings through a supergraph are determined by jointly learning architecture distribution parameters and neural operation parameters through backpropagation, using a novel resource-aware loss function. Combined, the routings form branching structures which exhibit high overall performance while being computationally efficient, as we demonstrate across several datasets (PASCAL-Context, NYUD-v2) and network backbones (ResNet-50, MobileNetV2). The proposed method is highly flexible and can serve as a basis for further exploration in MTL NAS.

\newpage

\begin{appendices}

\beginappendixa
\section{Training Settings}

In this section, we describe the training setup used for experiments on PASCAL-Context. On NYUD-v2, the exact same setup was used, except that the number of training iterations was halved. Training and evaluation code for this project was written using the {\tt PyTorch} library~\cite{paszke2017automatic}.

\textbf{Data augmentation.} We augment input images during training by random scaling with values between 0.5 and 2.0 (in increments of 0.25), random cropping to input size (which was fixed to $512 \times 512$ for full-resolution PASCAL-Context) and random horizontal flipping. Image intensities are rescaled to the [-1, 1] range.

\textbf{Task losses.} For semantic segmentation and human parts segmentation we use a cross-entropy loss (loss weights $\omega_t = 1$ and $\omega_t = 2$ in Eq.~\ref{eq:weighted_sum}, respectively), for saliency estimation a balanced cross-entropy loss ($\omega_t = 5$), for depth estimation a $\mathcal{L}_1$ loss ($\omega_t = 1$), for surface normal estimation a $\mathcal{L}_1$ loss with unit vector normalization ($\omega_t = 10$) and for edge detection a weighted cross-entropy loss ($\omega_t = 50$). For edge detection, the positive pixels are weighted with 0.95 and the negative pixels with 0.05 on PASCAL-Context, while on NYUD-v2 the weights are 0.8 and 0.2. $\omega_t$ for each task was found by conducting a logarithmic grid search over candidate values with single-task networks.

\textbf{Optimization hyperparameters.} Model weights $\theta$ are updated using Stochastic Gradient Descent (SGD) with momentum of 0.9 and weight decay 0.0001. The initial learning rate is set to 0.005 and decayed during training according to a `poly' learning rate policy~\cite{chen2017deeplab}. For the architecture distribution parameters $\alpha$, we use an Adam optimizer~\cite{kingma2014adam} with learning rate 0.01 and weight decay 0.00005. We use a batchsize of 8 and 16 for ResNet-50 and MobileNetV2, respectively.

\textbf{Architecture search.}  We update the supergraph sequentially for each task. Before the architecture search, we `warm up' the supergraph by training each operation's model weights $\theta$ (initialized with ImageNet weights) on the corresponding task only for 2000 iterations. The architecture distribution parameters $\alpha$ are initialized with zeros. During the search, we alternatively train $\alpha$ on 20\% of the data and $\theta$ on the other 80\%. This cycle is repeated until $\theta$ has received 40000 updates. Over the course of training, the Gumbel-Softmax temperature $\tau$ is annealed linearly from 5.0 to 0.1. Importantly, we use the batch-specific statistics for batch normalization during the $\alpha$ update phase and reset the batch statistics before training $\theta$ after every architecture change. Furthermore, to equalize the scale of candidate operations for the search, we disable learnable affine parameters in the last batch normalization of every operation. Finally, the momentum of the $\theta$-optimizer is reset after every change to the architecture. 

\textbf{Branched network training.} After the architecture search, the resulting branched network is retrained from scratch for 40000 iterations. The encoder network weights are initialized with ImageNet weights. For all operations that are shared between several tasks, we divide the learning rate by the number of tasks sharing, since those operations receive more updates.

\beginappendixb
\section{Implementation of Baselines}

For FAFS~\cite{lu2017fully}, we pre-train a fully shared network on all the tasks and then calculate the task groupings in all layers greedily (starting from the last) according to the task affinity measure described in~\cite{lu2017fully}. For edge detection, we use the loss instead of the optimal dataset F-measure to determine sample difficulty. Finding branching structures via the BMN approach~\cite{vandenhende2019branched} involved training separate single-task networks and computing the Representational Similarity Analysis matrix from the resulting task-specific feature maps, exactly as described in the paper. Various branching structures can be found by exhaustively searching candidates among a reduced pool, containing all possible structures below a specified MAdds value. To keep the comparison fair, we trained the branched structures resulting from BMN and FAFS in exactly the same setting as ours.

We implemented Cross-Stitch Networks~\cite{misra2016cross}, NDDR-CNN~\cite{gao2019nddr} and MTAN~\cite{liu2019end} based on the code provided by the authors and information given in the papers. We use a similar training setup as the one described for our method, however we conducted a logarithmic grid search over learning rates for each baseline individually. For Cross-Stitch Networks, applying one unit per feature tensor (as opposed to channel-wise) yielded more stable results. The weights of Cross-Stitch- and NDDR-CNN-units are initialized with $\alpha=0.8$ and $\beta=\frac{0.2}{T-1}$, where $T$ is the number of tasks. Both methods are applied on the fully pre-trained single-task networks. For MTAN with the MobileNetV2 backbone, we change the $3\times3$ convolutions in the attention modules to depthwise separable convolutions. In general, all ReLU activations are replaced with ReLU6 for MobileNetV2.

\beginappendixc
\section{Implementation Verification}

\begin{table}
\begin{center}
\resizebox{\linewidth}{!}{%
\ra{1.3}
\footnotesize
\begin{tabularx}{\textwidth}{@{}lYYYYr@{}}\toprule
Backbone & SemSeg\,$\uparrow$ & PartSeg\,$\uparrow$ & Sal\,$\uparrow$ & Norm\,$\downarrow$ & Edge\,$\uparrow$ \\ \midrule
MobileNetV2, \cite{maninis2019attentive} & 62.10 & 54.88 & 66.30 & 14.88 & 69.50 \\
MobileNetV2, ours & 65.11 & 57.54 & 65.41 & 13.98 & 69.50 \\
\hdashline\noalign{\vskip 0.5ex}
ResNet-50, \cite{maninis2019attentive} & 68.30 & 60.70 & 65.40 & 14.61 & 72.70 \\
ResNet-50, ours & 70.43 & 63.93 & 67.34 & 13.39 & 74.10 \\
\bottomrule
\end{tabularx}
}
\end{center}
\caption{DeepLabv3+ performance in a single-task setting on PASCAL-Context using either MobileNetV2 or ResNet-50 as a backbone. We compare the performance obtained using our implementation with the results published in~\cite{maninis2019attentive}.}
\label{tab:perform_comp}
\end{table}

To show that our implementations of DeepLabv3+ with the above mentioned backbones perform competitively for the tasks of interest, we compare in Table~\ref{tab:perform_comp} our single-task performances on PASCAL-Context with published results in~\cite{maninis2019attentive}. A direct comparison is inconclusive even though the architectures are analogous, as the results in~\cite{maninis2019attentive} are obtained with different training setups. Nevertheless, the numbers demonstrate that our single-task networks represent a strong baseline for comparison.

\beginappendixd
\section{Complementary Results}

In Table~\ref{tab:resnet_pascal}, we present the performances of our method and simple baselines for a ResNet-50 backbone on PASCAL-Context (plotted on the right in Fig.~\ref{fig:res}). For this setting, we choose not to report scores for Cross-Stitch Networks~\cite{misra2016cross}, NDDR-CNN~\cite{gao2019nddr} and MTAN~\cite{liu2019end} since we were unable to obtain competitive performances for those approaches, despite the extensive learning rate grid-search.

\begin{table}
\begin{center}
\resizebox{\linewidth}{!}{%
\ra{1.3}
\footnotesize
\begin{tabularx}{\textwidth}{@{}lYYYYYYr@{}}\toprule
Model & MAdds\,$\downarrow$ & SemSeg\,$\uparrow$ & PartSeg\,$\uparrow$ & Sal\,$\uparrow$ & Norm\,$\downarrow$ & Edge\,$\uparrow$ & $\Delta_m$ [\%]\,$\uparrow$ \\ \midrule
Single & 345.7B & 70.43 & 63.93 & 67.34 & 13.39 & 74.10 & 0.00 \\
Shared & 154.6B & 68.24 & 62.18 & 65.16 & 14.98 & 71.90 & -4.80 \\
\hdashline\noalign{\vskip 0.5ex}
BMTAS-1 & 199.0B & 68.17 & 62.36 & 65.64 & 14.09 & 72.20 & -3.20 \\
BMTAS-2 & 225.8B & 66.92 & 62.93 & 65.82 & 13.70 & 72.90 & -2.56 \\
BMTAS-3 & 298.2B & 69.58 & 64.36 & 66.68 & 13.65 & 73.00 & -1.00 \\
\bottomrule
\end{tabularx}
}
\end{center}
\caption{Comparison of our method with simple baselines on PASCAL-Context using a ResNet-50 backbone. The resource loss weights $\lambda$ for BMTAS are \{0.02, 0.005, 0.001\} respectively.}
\label{tab:resnet_pascal}
\end{table}

\end{appendices}

\bibliography{egbib}

\begin{thebibliography}{50}
\providecommand{\natexlab}[1]{#1}
\providecommand{\url}[1]{\texttt{#1}}
\expandafter\ifx\csname urlstyle\endcsname\relax
  \providecommand{\doi}[1]{doi: #1}\else
  \providecommand{\doi}{doi: \begingroup \urlstyle{rm}\Url}\fi

\bibitem[Bingel and S{\o}gaard(2017)]{bingel2017identifying}
Joachim Bingel and Anders S{\o}gaard.
\newblock Identifying beneficial task relations for multi-task learning in deep
  neural networks.
\newblock In \emph{European Chapter of the Association for Computational
  Linguistics}, 2017.

\bibitem[Bragman et~al.(2019)Bragman, Tanno, Ourselin, Alexander, and
  Cardoso]{bragman2019stochastic}
Felix~JS Bragman, Ryutaro Tanno, Sebastien Ourselin, Daniel~C Alexander, and
  Jorge Cardoso.
\newblock Stochastic filter groups for multi-task cnns: Learning specialist and
  generalist convolution kernels.
\newblock In \emph{ICCV}, 2019.

\bibitem[Cai et~al.(2019)Cai, Zhu, and Han]{cai2018proxylessnas}
Han Cai, Ligeng Zhu, and Song Han.
\newblock Proxylessnas: Direct neural architecture search on target task and
  hardware.
\newblock In \emph{ICLR}, 2019.

\bibitem[Chen et~al.(2017)Chen, Papandreou, Kokkinos, Murphy, and
  Yuille]{chen2017deeplab}
Liang-Chieh Chen, George Papandreou, Iasonas Kokkinos, Kevin Murphy, and Alan~L
  Yuille.
\newblock Deeplab: Semantic image segmentation with deep convolutional nets,
  atrous convolution, and fully connected crfs.
\newblock \emph{TPAMI}, 40\penalty0 (4):\penalty0 834--848, 2017.

\bibitem[Chen et~al.(2018{\natexlab{a}})Chen, Zhu, Papandreou, Schroff, and
  Adam]{chen2018encoder}
Liang-Chieh Chen, Yukun Zhu, George Papandreou, Florian Schroff, and Hartwig
  Adam.
\newblock Encoder-decoder with atrous separable convolution for semantic image
  segmentation.
\newblock In \emph{ECCV}, 2018{\natexlab{a}}.

\bibitem[Chen et~al.(2014)Chen, Mottaghi, Liu, Fidler, Urtasun, and
  Yuille]{chen2014detect}
Xianjie Chen, Roozbeh Mottaghi, Xiaobai Liu, Sanja Fidler, Raquel Urtasun, and
  Alan Yuille.
\newblock Detect what you can: Detecting and representing objects using
  holistic models and body parts.
\newblock In \emph{CVPR}, 2014.

\bibitem[Chen et~al.(2018{\natexlab{b}})Chen, Badrinarayanan, Lee, and
  Rabinovich]{chen2017gradnorm}
Zhao Chen, Vijay Badrinarayanan, Chen-Yu Lee, and Andrew Rabinovich.
\newblock Gradnorm: Gradient normalization for adaptive loss balancing in deep
  multitask networks.
\newblock In \emph{ICML}, 2018{\natexlab{b}}.

\bibitem[Everingham et~al.(2010)Everingham, Van~Gool, Williams, Winn, and
  Zisserman]{everingham2010pascal}
Mark Everingham, Luc Van~Gool, Christopher~KI Williams, John Winn, and Andrew
  Zisserman.
\newblock The pascal visual object classes (voc) challenge.
\newblock \emph{IJCV}, 88\penalty0 (2):\penalty0 303--338, 2010.

\bibitem[Gao et~al.(2019)Gao, Ma, Zhao, Liu, and Yuille]{gao2019nddr}
Yuan Gao, Jiayi Ma, Mingbo Zhao, Wei Liu, and Alan~L Yuille.
\newblock Nddr-cnn: Layerwise feature fusing in multi-task cnns by neural
  discriminative dimensionality reduction.
\newblock In \emph{CVPR}, 2019.

\bibitem[Gao et~al.(2020)Gao, Bai, Jie, Ma, Jia, and Liu]{gao2020mtl}
Yuan Gao, Haoping Bai, Zequn Jie, Jiayi Ma, Kui Jia, and Wei Liu.
\newblock Mtl-nas: Task-agnostic neural architecture search towards
  general-purpose multi-task learning.
\newblock In \emph{CVPR}, 2020.

\bibitem[Girshick et~al.(2014)Girshick, Donahue, Darrell, and
  Malik]{girshick2014rich}
Ross Girshick, Jeff Donahue, Trevor Darrell, and Jitendra Malik.
\newblock Rich feature hierarchies for accurate object detection and semantic
  segmentation.
\newblock In \emph{CVPR}, 2014.

\bibitem[Gordon et~al.(2018)Gordon, Eban, Nachum, Chen, Wu, Yang, and
  Choi]{gordon2018morphnet}
Ariel Gordon, Elad Eban, Ofir Nachum, Bo~Chen, Hao Wu, Tien-Ju Yang, and Edward
  Choi.
\newblock Morphnet: Fast \& simple resource-constrained structure learning of
  deep networks.
\newblock In \emph{CVPR}, 2018.

\bibitem[Guo et~al.(2018)Guo, Haque, Huang, Yeung, and Fei-Fei]{guo2018dynamic}
Michelle Guo, Albert Haque, De-An Huang, Serena Yeung, and Li~Fei-Fei.
\newblock Dynamic task prioritization for multitask learning.
\newblock In \emph{ECCV}, 2018.

\bibitem[He et~al.(2016)He, Zhang, Ren, and Sun]{he2016deep}
Kaiming He, Xiangyu Zhang, Shaoqing Ren, and Jian Sun.
\newblock Deep residual learning for image recognition.
\newblock In \emph{CVPR}, 2016.

\bibitem[Jang et~al.(2017)Jang, Gu, and Poole]{jang2016categorical}
Eric Jang, Shixiang Gu, and Ben Poole.
\newblock Categorical reparameterization with gumbel-softmax.
\newblock In \emph{ICLR}, 2017.

\bibitem[Kanakis et~al.(2020)Kanakis, Bruggemann, Saha, Georgoulis, Obukhov,
  and Van~Gool]{kanakis2020reparameterizing}
Menelaos Kanakis, David Bruggemann, Suman Saha, Stamatios Georgoulis, Anton
  Obukhov, and Luc Van~Gool.
\newblock Reparameterizing convolutions for incremental multi-task learning
  without task interference.
\newblock In \emph{ECCV}, 2020.

\bibitem[Kendall et~al.(2018)Kendall, Gal, and Cipolla]{kendall2018multi}
Alex Kendall, Yarin Gal, and Roberto Cipolla.
\newblock Multi-task learning using uncertainty to weigh losses for scene
  geometry and semantics.
\newblock In \emph{CVPR}, 2018.

\bibitem[Kingma and Ba(2015)]{kingma2014adam}
Diederik~P Kingma and Jimmy Ba.
\newblock Adam: A method for stochastic optimization.
\newblock In \emph{ICLR}, 2015.

\bibitem[Kokkinos(2017)]{kokkinos2017ubernet}
Iasonas Kokkinos.
\newblock Ubernet: Training a universal convolutional neural network for low-,
  mid-, and high-level vision using diverse datasets and limited memory.
\newblock In \emph{CVPR}, 2017.

\bibitem[Liu et~al.(2019{\natexlab{a}})Liu, Chen, Schroff, Adam, Hua, Yuille,
  and Fei-Fei]{liu2019auto}
Chenxi Liu, Liang-Chieh Chen, Florian Schroff, Hartwig Adam, Wei Hua, Alan~L
  Yuille, and Li~Fei-Fei.
\newblock Auto-deeplab: Hierarchical neural architecture search for semantic
  image segmentation.
\newblock In \emph{CVPR}, 2019{\natexlab{a}}.

\bibitem[Liu et~al.(2019{\natexlab{b}})Liu, Simonyan, and Yang]{liu2018darts}
Hanxiao Liu, Karen Simonyan, and Yiming Yang.
\newblock Darts: Differentiable architecture search.
\newblock In \emph{ICLR}, 2019{\natexlab{b}}.

\bibitem[Liu et~al.(2019{\natexlab{c}})Liu, Johns, and Davison]{liu2019end}
Shikun Liu, Edward Johns, and Andrew~J Davison.
\newblock End-to-end multi-task learning with attention.
\newblock In \emph{CVPR}, 2019{\natexlab{c}}.

\bibitem[Long et~al.(2015)Long, Shelhamer, and Darrell]{long2015fully}
Jonathan Long, Evan Shelhamer, and Trevor Darrell.
\newblock Fully convolutional networks for semantic segmentation.
\newblock In \emph{CVPR}, 2015.

\bibitem[Long et~al.(2017)Long, Cao, Wang, and Philip]{long2017learning}
Mingsheng Long, Zhangjie Cao, Jianmin Wang, and S~Yu Philip.
\newblock Learning multiple tasks with multilinear relationship networks.
\newblock In \emph{NeurIPS}, 2017.

\bibitem[Lu et~al.(2017)Lu, Kumar, Zhai, Cheng, Javidi, and Feris]{lu2017fully}
Yongxi Lu, Abhishek Kumar, Shuangfei Zhai, Yu~Cheng, Tara Javidi, and Rogerio
  Feris.
\newblock Fully-adaptive feature sharing in multi-task networks with
  applications in person attribute classification.
\newblock In \emph{CVPR}, 2017.

\bibitem[Maddison et~al.(2017)Maddison, Mnih, and Teh]{maddison2016concrete}
Chris~J Maddison, Andriy Mnih, and Yee~Whye Teh.
\newblock The concrete distribution: A continuous relaxation of discrete random
  variables.
\newblock In \emph{ICLR}, 2017.

\bibitem[Maninis et~al.(2019)Maninis, Radosavovic, and
  Kokkinos]{maninis2019attentive}
Kevis-Kokitsi Maninis, Ilija Radosavovic, and Iasonas Kokkinos.
\newblock Attentive single-tasking of multiple tasks.
\newblock In \emph{CVPR}, 2019.

\bibitem[Martin et~al.(2004)Martin, Fowlkes, and Malik]{martin2004learning}
David~R Martin, Charless~C Fowlkes, and Jitendra Malik.
\newblock Learning to detect natural image boundaries using local brightness,
  color, and texture cues.
\newblock \emph{TPAMI}, 26\penalty0 (5):\penalty0 530--549, 2004.

\bibitem[Misra et~al.(2016)Misra, Shrivastava, Gupta, and
  Hebert]{misra2016cross}
Ishan Misra, Abhinav Shrivastava, Abhinav Gupta, and Martial Hebert.
\newblock Cross-stitch networks for multi-task learning.
\newblock In \emph{CVPR}, 2016.

\bibitem[Molchanov et~al.(2017)Molchanov, Tyree, Karras, Aila, and
  Kautz]{molchanov2016pruning}
Pavlo Molchanov, Stephen Tyree, Tero Karras, Timo Aila, and Jan Kautz.
\newblock Pruning convolutional neural networks for resource efficient
  inference.
\newblock In \emph{ICLR}, 2017.

\bibitem[Neven et~al.(2017)Neven, De~Brabandere, Georgoulis, Proesmans, and
  Van~Gool]{neven2017fast}
Davy Neven, Bert De~Brabandere, Stamatios Georgoulis, Marc Proesmans, and Luc
  Van~Gool.
\newblock Fast scene understanding for autonomous driving.
\newblock In \emph{IEEE Symposium on Intelligent Vehicles Workshop}, 2017.

\bibitem[Paszke et~al.(2017)Paszke, Gross, Chintala, Chanan, Yang, DeVito, Lin,
  Desmaison, Antiga, and Lerer]{paszke2017automatic}
Adam Paszke, Sam Gross, Soumith Chintala, Gregory Chanan, Edward Yang, Zachary
  DeVito, Zeming Lin, Alban Desmaison, Luca Antiga, and Adam Lerer.
\newblock Automatic differentiation in pytorch.
\newblock In \emph{NeurIPS Workshop}, 2017.

\bibitem[Pham et~al.(2018)Pham, Guan, Zoph, Le, and Dean]{pham2018efficient}
Hieu Pham, Melody~Y Guan, Barret Zoph, Quoc~V Le, and Jeff Dean.
\newblock Efficient neural architecture search via parameter sharing.
\newblock In \emph{ICML}, 2018.

\bibitem[Real et~al.(2017)Real, Moore, Selle, Saxena, Suematsu, Tan, Le, and
  Kurakin]{real2017large}
Esteban Real, Sherry Moore, Andrew Selle, Saurabh Saxena, Yutaka~Leon Suematsu,
  Jie Tan, Quoc~V Le, and Alexey Kurakin.
\newblock Large-scale evolution of image classifiers.
\newblock In \emph{ICML}, 2017.

\bibitem[Real et~al.(2019)Real, Aggarwal, Huang, and Le]{real2019regularized}
Esteban Real, Alok Aggarwal, Yanping Huang, and Quoc~V Le.
\newblock Regularized evolution for image classifier architecture search.
\newblock In \emph{AAAI}, 2019.

\bibitem[Rosenbaum et~al.(2018)Rosenbaum, Klinger, and
  Riemer]{rosenbaum2017routing}
Clemens Rosenbaum, Tim Klinger, and Matthew Riemer.
\newblock Routing networks: Adaptive selection of non-linear functions for
  multi-task learning.
\newblock In \emph{ICLR}, 2018.

\bibitem[Ruder(2017)]{ruder2017overview}
Sebastian Ruder.
\newblock An overview of multi-task learning in deep neural networks.
\newblock \emph{arXiv preprint arXiv:1706.05098}, 2017.

\bibitem[Ruder et~al.(2019)Ruder, Bingel, Augenstein, and
  S{\o}gaard]{ruder2019latent}
Sebastian Ruder, Joachim Bingel, Isabelle Augenstein, and Anders S{\o}gaard.
\newblock Latent multi-task architecture learning.
\newblock In \emph{AAAI}, 2019.

\bibitem[Sandler et~al.(2018)Sandler, Howard, Zhu, Zhmoginov, and
  Chen]{sandler2018mobilenetv2}
Mark Sandler, Andrew Howard, Menglong Zhu, Andrey Zhmoginov, and Liang-Chieh
  Chen.
\newblock Mobilenetv2: Inverted residuals and linear bottlenecks.
\newblock In \emph{CVPR}, 2018.

\bibitem[Sener and Koltun(2018)]{sener2018multi}
Ozan Sener and Vladlen Koltun.
\newblock Multi-task learning as multi-objective optimization.
\newblock In \emph{NeurIPS}, 2018.

\bibitem[Silberman et~al.(2012)Silberman, Hoiem, Kohli, and
  Fergus]{silberman2012indoor}
Nathan Silberman, Derek Hoiem, Pushmeet Kohli, and Rob Fergus.
\newblock Indoor segmentation and support inference from rgbd images.
\newblock In \emph{ECCV}, 2012.

\bibitem[Vandenhende et~al.(2020{\natexlab{a}})Vandenhende, Georgoulis,
  De~Brabandere, and Van~Gool]{vandenhende2019branched}
Simon Vandenhende, Stamatios Georgoulis, Bert De~Brabandere, and Luc Van~Gool.
\newblock Branched multi-task networks: deciding what layers to share.
\newblock In \emph{BMVC}, 2020{\natexlab{a}}.

\bibitem[Vandenhende et~al.(2020{\natexlab{b}})Vandenhende, Georgoulis, and
  Van~Gool]{vandenhende2020mti}
Simon Vandenhende, Stamatios Georgoulis, and Luc Van~Gool.
\newblock Mti-net: Multi-scale task interaction networks for multi-task
  learning.
\newblock In \emph{ECCV}, 2020{\natexlab{b}}.

\bibitem[Veniat and Denoyer(2018)]{veniat2018learning}
Tom Veniat and Ludovic Denoyer.
\newblock Learning time/memory-efficient deep architectures with budgeted super
  networks.
\newblock In \emph{CVPR}, 2018.

\bibitem[Wu et~al.(2019)Wu, Dai, Zhang, Wang, Sun, Wu, Tian, Vajda, Jia, and
  Keutzer]{wu2019fbnet}
Bichen Wu, Xiaoliang Dai, Peizhao Zhang, Yanghan Wang, Fei Sun, Yiming Wu,
  Yuandong Tian, Peter Vajda, Yangqing Jia, and Kurt Keutzer.
\newblock Fbnet: Hardware-aware efficient convnet design via differentiable
  neural architecture search.
\newblock In \emph{CVPR}, 2019.

\bibitem[Xie and Tu(2015)]{xie2015holistically}
Saining Xie and Zhuowen Tu.
\newblock Holistically-nested edge detection.
\newblock In \emph{ICCV}, 2015.

\bibitem[Xie et~al.(2019)Xie, Zheng, Liu, and Lin]{xie2018snas}
Sirui Xie, Hehui Zheng, Chunxiao Liu, and Liang Lin.
\newblock Snas: stochastic neural architecture search.
\newblock In \emph{ICLR}, 2019.

\bibitem[Xu et~al.(2018)Xu, Ouyang, Wang, and Sebe]{xu2018pad}
Dan Xu, Wanli Ouyang, Xiaogang Wang, and Nicu Sebe.
\newblock Pad-net: Multi-tasks guided prediction-and-distillation network for
  simultaneous depth estimation and scene parsing.
\newblock In \emph{CVPR}, 2018.

\bibitem[Zoph and Le(2017)]{zoph2016neural}
Barret Zoph and Quoc~V Le.
\newblock Neural architecture search with reinforcement learning.
\newblock In \emph{ICLR}, 2017.

\bibitem[Zoph et~al.(2018)Zoph, Vasudevan, Shlens, and Le]{zoph2018learning}
Barret Zoph, Vijay Vasudevan, Jonathon Shlens, and Quoc~V Le.
\newblock Learning transferable architectures for scalable image recognition.
\newblock In \emph{CVPR}, 2018.

\end{thebibliography}
\end{document}